\pdfoutput=1
\documentclass[11pt,a4paper]{article}
\usepackage{authblk}
\usepackage[nohyperref]{naaclhlt2019}
\usepackage{times}
\usepackage{latexsym}
\usepackage{booktabs}
\usepackage{url}
\usepackage{graphicx}
\usepackage{amsmath}

\aclfinalcopy % Uncomment this line for the final submission
\title{Legal Area Classification: A Comparative Study of Text Classifiers on Singapore Supreme Court Judgments}

\author[*]{\bf Jerrold Soh Tsin Howe}
\author[ ]{\bf Lim How Khang}
\author[**]{\bf Ian Ernst Chai}
\affil[*]{Singapore Management University, School of Law}
\affil[**]{Attorney-General's Chambers, Singapore}
\affil [ ]{\tt jerroldsoh@smu.edu.sg, howkhang.lim@gmail.com}
\affil[ ]{\tt ian\_ernst\_chai@agc.gov.sg}

\date{}

\begin{document}

\maketitle

\begin{abstract}
This paper conducts a comparative study on the performance of various machine learning (``ML'') approaches for classifying judgments into legal areas. Using a novel dataset of 6,227 Singapore Supreme Court judgments, we investigate how state-of-the-art NLP methods compare against traditional statistical models when applied to a legal corpus that comprised few but lengthy documents. All approaches tested, including topic model, word embedding, and language model-based classifiers, performed well with as little as a few hundred judgments. However, more work needs to be done to optimize state-of-the-art methods for the legal domain.

\end{abstract}

\section{Introduction}
\label{sec:intro}

Every legal case falls into one or more areas of law (``legal areas''). These areas are lawyers' shorthand for the subset of legal principles and rules governing the case. Thus lawyers often triage a new case by asking if it falls within tort, contract, or other legal areas. Answering this allows unresolved cases to be funneled to the right experts and for resolved precedents to be efficiently retrieved. Legal database providers routinely provide area-based search functionality; courts often publish judgments labelled by legal area.

The law therefore yields pockets of expert-labelled text. A system that classifies legal texts by area would be useful for enriching older, typically unlabelled judgments with metadata for more efficient search and retrieval. The system can also suggest areas for further inquiry by predicting which areas a new text falls within.

Despite its potential, this problem, which we refer to and later define as ``legal area classification'', remains relatively unexplored. One explanation is the relative scarcity of labelled documents in the law (typically in the low thousands), at least by deep learning standards. This problem is acute in smaller jurisdictions like Singapore, where the number of labelled cases is limited by the few cases that actually reach the courts. Another explanation is that legal texts are typically longer than the customer reviews, tweets, and other documents typical in NLP research.

Against this backdrop, this paper uses a novel dataset of Singapore Supreme Court judgments to comparatively study the performance of various text classification approaches for legal area classification. Our specific research question is as follows: how do recent state-of-the-art models compare against traditional statistical models when applied to legal corpora that, typically, comprise few but lengthy documents?

We find that there are challenges when it comes to adapting state-of-the-art deep learning classifiers for tasks in the legal domain. Traditional topic models still outperform the more recent neural-based classifiers on certain metrics, suggesting that emerging research (fit specially to tasks with numerous short documents) may not carry well into the legal domain unless more work is done to optimize them for legal NLP tasks. However, that shallow models perform well suggests that enough exploitable information exists in legal texts for deep learning approaches better-tailored to the legal domain to perform as well if not better.

\section{Related Work}
\label{sec:related_work}

Papers closest to ours are those that likewise examine legal area classification. \citet{goncalves_2005} used bag-of-words (``BOW'') features learned using TF-IDF to train linear support vector machines (``linSVMs'') to classify decisions of the Portuguese Attorney General's Office into 10 legal areas. \citet{eunomos_2012} used TF-IDF features enriched by a semi-automatically linked legal ontology and linSVMs to classify Italian legislation into 15 civil law areas. \citet{sulea_zampieri_vela_genabith_2017} classified French Supreme Court judgments into 8 civil law areas, again using BOW features learned using Latent Semantic Analysis (``LSA'') \cite{scott_et_al_1990} and linSVMs.

On legal text classification more generally, \citet{aletras_2016, liu_chen_2017, sulea_zampieri_vela_genabith_2017} used BOW features extracted from judgments and linSVMs for predicting case outcomes. \citet{kellner_2012} use BOW features and a linSVM to classify contract clauses. 

NLP has also been used for legal information extraction. \citet{venkatesh_2013} used Latent Dirichlet Allocation \cite{blei_ng_jordan_2003} (``LDA'') to cluster Indian court judgments. \citet{falakmasir_2017} used vector space models to extract legal factors motivating case outcomes from American trade secret misappropriation judgments.

There is also growing scholarship on legal text analysis. Typically, topic models are used to extract $N$-gram clusters from legal corpora, such as Constitutions, statutes, and Parliamentary records, then assessed for legal significance \cite{young_2013, carter_brown_rahmani_2018}. More recently, \citet{chen_ash_2018} used document embeddings trained on United States Supreme Court judgments to encode and study spatial and temporal patterns across federal judges and appellate courts.

We contribute to this literature by (1) benchmarking new text classification techniques against legal area classification, and (2) more deeply exploring how document scarcity and length affect performance. Beyond BOW features and linSVMs, we use word embeddings and newly-developed language models. Our novel label set comprises 31 legal areas relevant to Singapore's \textit{common} law system. Judgments of the Singapore Supreme Court have thus far not been exploited. We also draw an important but overlooked distinction between \textit{cases} and \textit{judgments}.

\section{Problem Description}
\label{sec:task}

Legal areas generally refer to a subset of related legal principles and rules governing certain dispute types. There is no universal set of legal areas. Areas like tort and equity, well-known in English and American law, have no direct analogue in certain civil law systems. Societal change may create new areas of law like data protection. However, the set of legal areas in a given jurisdiction and time is well-defined. Denote this as $L$.

Lawyers typically attribute a given case $c_i$ to a given legal area $l \in L$ if $c_i$'s attributes $v_{c_i}$ (\textit{e.g.} a vector of its facts, parties involved, and procedural trail) raise legal issues that implicate some principle or rule in $l$. Cases may fall into more than one legal area but never none.

Cases should be distinguished from the \textit{judgments} courts write when resolving them (denoted $j_{c_i}$). $j_{c_i}$ may not state everything in $v_{c_i}$ because judges need only discuss issues material to how the case should be resolved. Suppose a claimant mounts two claims on the same issue against a defendant in tort, and in trademark law. If the judge finds for the claimant in tort, he/she may not discuss trademark at all (though some may still do so). Thus, even though $v_{c_i}$ raises trademark issues, $j_{c_i}$ may not contain any $N$-grams discussing the same. It is possible that a  $v_{c_i}$ we would assign to $l$ leads to a $j_{c_i}$ we would not assign to $l$. The upshot is that judgments are incomplete sources of case information; classifying \textit{judgments} is not the same as classifying \textit{cases}.

This paper focuses on the former. We treat this as a supervised legal text multi-class and multi-label classification task. The goal is to learn $f^*: j_i \mapsto L_{j_i}$ where $^*$ denotes optimality.

\section{Data}
\label{sec:data}

The corpus comprises 6,227 judgments of the Singapore Supreme Court written in English.\footnote{The judgments were issued between 3 January 2000 and 18 February 2019 and were downloaded from \url{http://www.singaporelawwatch.sg}, an official repository of Singapore Supreme Court judgments.} Each judgment comes in PDF format, with its legal areas labelled by the Court. The median judgment has 6,968 tokens and is significantly longer than the typical customer review or news article commonly found in datasets for benchmarking machine learning models on text classification. 

The raw dataset yielded 51 different legal area labels. Some  labels were subsets of larger legal areas and were manually merged into those. Label imbalance was present in the dataset so we limited the label set to the 30 most frequent areas. Remaining labels (252 in total) were then mapped to the label ``others''. Table \ref{label_dist} shows the final label distribution, truncated for brevity. Appendix \ref{app:label_map} presents the full label distribution and all label merging decisions.

\begin{table}[t!]
\begin{center}
\begin{tabular}{lr}
\toprule 
\bf Label & \bf Count \\
\midrule
civil\_procedure                   &           1369 \\
contract\_law                      &            861 \\
criminal\_procedure\_and\_sentencing &            775 \\
criminal\_law                      &            734 \\
family\_law                        &            491 \\
...                               &            ... \\
others                            &            252 \\
...                               &            ... \\
banking\_law                       &             75 \\
restitution                       &             60 \\
agency\_law                        &             57 \\
res\_judicata                      &             49 \\
insurance\_law                     &             39 \\
\bf Total                           &        \bf 8853 \\
\bottomrule
\end{tabular}
\end{center}
\caption{\label{label_dist} Truncated Distribution of Final Labels}
\end{table}

\section{Models and Methods}

Given label imbalance, we held out 10\% of the corpus by stratified iterative sampling \cite{it-strat1, it-strat2}. For each model type, we trained three separate classifiers on \textit{the same} 10\% (n=588), 50\% (n=2795), and 100\% (n=5599) of the remaining \textit{training} set (``training subsets''), again split by stratified iteration, and tested them against \textit{the same} 10\% holdout. We studied four model types of increasing sophistication and recency. These are briefly explained here. Further implementation details may be found in the Appendix \ref{app:implementation}.

\subsection{Baseline Models}

$base_{pdf}$ is a dummy classifier which predicts 1 for any label which expectation equals or exceeds $1 / 31$ (the total number of labels).

$count_m$ uses a keyword matching strategy that emulates how lawyers may approach the task. It predicts 1 for any label if its associated terms appear $\geq m$ non-unique times in $j_i$. $m$ is a manually-set threshold. A label's set of associated terms is the union of (a) the set of its sub-labels in the training subset, and (b) the set of non-stopword unigrams in the label itself. We manually added potentially problematic unigrams like ``law'' to the stopwords list. Suppose the label ``tort\_law'' appears twice in the training subset, first with sub-label ``negligence'', and later with sub-label ``harassment''. The associated terms set would be $\{tort, negligence, harassment\}$.

\subsection{Topic Models}

$lsa_k$ is a one-vs-rest linSVM trained using $k$ topics extracted by LSA. We used LSA and linSVMs as benchmarks because, despite their vintage, they remain a staple of the \textit{legal} text classification literature (see Section \ref{sec:related_work} above). Indeed, LDA models were also tested but strictly underperformed LSA models in all experiment and were thus not reported. Feature vectorizers and classifiers from scikit-learn \cite{sklearn} were re-trained for each training subset with all default settings except sublinear term frequencies were used in the TF-IDF step as recommended by \citet{sklearn_lsa}.

\subsection{Word Embedding Feature Models}

% Word embeddings represent words using dense vectors that have been learned through observing the contexts of each word within a large amount of unstructured text. 
Word vectors pre-trained on large corpora have been shown to capture syntactic and semantic word properties \cite{mikolov2013neurips,pennington2014glove}. We leverage on this by initializing word vectors using pre-trained GloVe vectors of length 300.\footnote{\url{http://nlp.stanford.edu/data/glove.6B.zip}} Judgment vectors were then composed in three ways: $glove_{avg}$ average-pools each word vector in $j_i$ (i.e.\ average-pooling); $glove_{max}$ uses max-pooling \cite{shen2018swem};  $glove_{cnn}$ feeds the word vectors through a shallow convolutional neural network (``CNN'') \cite{kim2014emnlp}. We chose to implement a shallow CNN model for $glove_{cnn}$ because it has been shown that deep CNN models do not necessarily perform better on text classification tasks \cite{le2018aaai}. To derive label predictions, judgment vectors were then fed through a multi-layer perceptron followed by a sigmoid function.

\subsection{Pre-trained Language Models}

Recent work has also shown that language representation models pre-trained on large unlabelled corpora and fine-tuned onto specific tasks significantly outperform models trained only on task-specific data. This method of transfer learning is particularly useful in legal NLP, given the lack of labelled data in the legal domain. We thus evaluated \citet{devlin2019bert}'s state-of-the-art BERT model using published pre-trained weights from $bert_{base}$ (12-layers; 110M parameters) and $bert_{large}$ (24-layers; 340M parameters).\footnote{\url{https://github.com/google-research/bert}} However, as BERT's self-attention transformer architecture \cite{vaswani2018neurips} only accepts up to 512 Wordpiece tokens \cite{wu2016wordpiece} as input, we used only the first 512 tokens of each $j_i$ to fine-tune both models.\footnote{Alternative strategies for selecting the 512 tokens trialed performed consistently worse and are not reported.} We considered splitting the judgment into shorter segments and passing each segment through the BERT model but doing so would require extensive modification to the original fine-tuning method; hence we left this for future experimentation. In this light, we also benchmarked \citet{howard2018ulmfit}'s ULMFiT model which accepts longer inputs due to its stacked-LSTM architecture.

\section{Results}

%\subsection{Comparison with Baselines}
Given our multi-label setting, we evaluated the models on and report micro- and macro-averaged F1 scores (Table \ref{results_f1}), precision (Table \ref{results_pr}), and recall (Table \ref{results_rec}). Micro-averaging calculates the metric globally while macro-averaging first calculates the metric \textit{within each label} before averaging across labels. Thus, micro-averaged metrics equally-weight each \textit{sample} and better indicate a model's performance on common labels whereas macro-averaged metrics equally-weight each \textit{label} and better indicate performance on rare labels.

%All models recorded significant gains above the statistical label-counting baseline $base_{pdf}$. This included the rules-based baseline $count_{25}$ which, to recall, emulates how lawyers may use keyword searches for this task. All ML classifiers outperformed $count_{25}$ by at least 5 percentage points even at the 10\% subset. As more training data was made available at the 50\% and 100\% subsets, this advantage widened to around 30 percentage points on average. This suggests that ML approaches are useful even on small legal corpora.
\subsection{F1 Score}

\begin{table}[ht!]
\begin{center}
\resizebox{\linewidth}{!}{
\begin{tabular}{lccc}
\toprule
\textbf{Subset} &           \textbf{10\%} &           \textbf{50\%} &          \textbf{100\%} \\
\midrule
$bert_{large}$ &  45.1 [57.9] &  56.7 [63.8] &  60.7 [66.3] \\
$bert_{base}$  &  43.1 [53.6] &  52.0 [57.6] &  56.2 [63.9] \\
$ulmfit$       &  \textbf{45.7} [62.8] &  45.9 [63.0] &  49.2 [64.3] \\
$glove_{cnn}$  &  40.7 [62.2] &  58.7 [67.1] &  63.1 [70.8] \\
$glove_{avg}$  &  36.7 [49.7] &  \textbf{59.1} [64.3] &  61.5 [65.6] \\
$glove_{max}$  &  29.2 [47.4] &  47.8 [59.9] &  52.5 [63.2] \\
$lsa_{250}$    &  37.9 [\textbf{63.5}] &  55.2 [\textbf{70.8}] &  \textbf{63.2} [\textbf{73.3}] \\
$lsa_{100}$    &  30.6 [58.5] &  51.8 [68.5] &  57.1 [70.8] \\
\midrule
$count_{25}$   &  32.6 [36.1] &  31.8 [30.6] &  27.7 [28.1] \\
$base_{pdf}$   &   5.2 [17.3] &   5.5 [16.6] &   5.5 [16.6] \\
\bottomrule
\end{tabular}}
\end{center}
\caption{\label{results_f1} Macro [Micro] F1 Scores Across Experiments}
\end{table}
% \addtolength{\tabcolsep}{0.75pt}

Across the three data subsets, all ML models consistently outperformed the statistical and keyword-matching baselines $base_{pdf}$ and $count_{25}$ respectively. Notably, even with limited training data (in the 10\% subset), most ML approaches surpassed $count_{25}$ which, to recall, emulates how lawyers may use keyword searches for legal area classification. Deep transfer learning approaches in particular performed well in this data-constrained setting, with $bert_{large}$, $bert_{base}$, and $ulmfit$ producing the best three macro-F1s. $ulmfit$ also achieved the second best micro-F1.

% except $glove_{max}$ and $lsa{100}$, which performed worse on the macro-averaged F1 score. 

As more training data became available at the 50\% and 100\% subsets, the ML classifiers' advantage over the baseline models widened to around 30 percentage points on average. Word-embedding models in particular showed significant improvements. $glove_{avg}$ and $glove_{cnn}$ outperformed most of the other models (with F1 scores of 63.1 and 61.5 respectively). Within the embedding models, $glove_{cnn}$ generally outperformed $glove_{avg}$ while $glove_{max}$ performed significantly worse than both and thus appears to be an unsuitable pooling strategy for this task.

Most surprisingly, $lsa_{250}$ emerged as the best performing model on both micro- and macro-averaged F1 scores for the 100\% subset. The model also produced the highest micro-averaged F1 score across all three data subsets, suggesting that common labels were handled well. $lsa_{250}$'s strong performance was fuelled primarily by high \textit{precision} rather than recall, as discussed below. 

\subsection{Precision}

\begin{table}[ht!]
\begin{center}
\resizebox{\linewidth}{!}{
\begin{tabular}{lccc}
\toprule
\textbf{Subset} &         \textbf{10\%} &         \textbf{50\%} &        \textbf{100\%} \\
\midrule
$bert_{large}$ &  54.7 [65.8] &  57.1 [59.7] &  63.6 [64.3] \\
$bert_{base}$  &  41.4 [45.1] &  48.1 [50.0] &  61.4 [67.2] \\
$ulmfit$       &  49.3 [63.7] &  46.6 [61.4] &  48.7 [63.2] \\
$glove_{cnn}$  &  50.7 [69.8] &  63.4 [68.5] &  66.7 [72.9] \\
$glove_{avg}$  &  \textbf{62.5} [68.0] &  67.0 [68.1] &  64.8 [68.2] \\
$glove_{max}$  &  51.3 [65.1] &  47.3 [56.6] &  59.2 [68.6] \\
$lsa_{250}$    &  56.7 [76.1] &  70.0 [81.1] &  \textbf{83.4} [81.7] \\
$lsa_{100}$    &  52.3 [\textbf{77.2}] &  \textbf{73.8} [\textbf{81.9}] &  73.9 [\textbf{83.7}] \\
\midrule
$count_{25}$   &  30.2 [26.4] &  26.4 [19.8] &  23.0 [17.8] \\
$base_{pdf}$   &   2.9 [10.0] &    3.1 [9.5] &    3.1 [9.5] \\
\bottomrule
\end{tabular}}
\end{center}
\caption{\label{results_pr} Macro [Micro] Precision Across Experiments}
\end{table}

As with F1 score, ML models outperformed baselines by large margins on precision. LSA models performed remarkably well here: except in the 10\% subset, where $glove_{cnn}$ recorded the highest macro-precision, top results for both precision measures belonged to either $lsa_{100}$ or $lsa_{250}$. Notably, on the 100\% subset, $lsa_{250}$ managed over 80\% on micro- and macro-precision.

% while $lsa_{100}$ had the highest micro-averaged precision. On the 50\% subset, the $lsa_{100}$ model had the highest precision scores, both micro and macro-averaged. When trained on the entire dataset, $lsa_{250}$ performed the best on macro-averaged precision while $lsa_{100}$ had the highest micro-averaged precision.

\subsection{Recall}

\begin{table}[ht!]
\begin{center}
\resizebox{\linewidth}{!}{
\begin{tabular}{lccc}
\toprule
\textbf{Subset} &         \textbf{10\%} &         \textbf{50\%} &        \textbf{100\%} \\
\midrule
$bert_{large}$ &  43.2 [51.7] &  \textbf{59.0} [\textbf{68.5}] &  61.6 [68.5] \\
$bert_{base}$  &  \textbf{50.0} [\textbf{66.1}] &  58.7 [67.9] &  54.2 [60.9] \\
$ulmfit$       &  46.1 [61.8] &  48.4 [64.8] &  52.6 [65.4] \\
$glove_{cnn}$  &  37.4 [56.0] &  58.2 [65.8] &  \textbf{62.3} [\textbf{68.8}] \\
$glove_{avg}$  &  28.7 [39.2] &  56.5 [60.9] &  62.1 [63.1] \\
$glove_{max}$  &  23.2 [37.2] &  49.9 [63.6] &  49.0 [58.5] \\
$lsa_{250}$    &  32.7 [54.4] &  50.2 [62.8] &  57.8 [66.5] \\
$lsa_{100}$    &  25.7 [47.0] &  45.3 [58.9] &  51.6 [61.4] \\
\midrule
$count_{25}$   &  48.1 [56.9] &  57.5 [66.3] &  59.9 [66.9] \\
$base_{pdf}$   &  29.0 [64.5] &  32.3 [67.7] &  32.3 [67.7] \\
\bottomrule
\end{tabular}}
\end{center}
\caption{\label{results_rec} Macro [Micro] Recall Across Experiments}
\end{table}

LSA's impressive results, however, stop short at recall. A striking observation from Table \ref{results_rec} is that LSA and most other ML models did \textit{worse} than $count_{25}$ on \textit{both} micro- and macro-recall \textit{across all data subsets}. Thus, a keyword-search strategy seems to be a simple yet strong baseline for identifying and retrieving judgments by legal area, particularly when recall is paramount and an ontology of area-related terms is available. To some extent this reflects realities in legal practice, where false negatives (missing relevant precedents) have greater potential to undermine legal argument than false positives (discovering irrelevant precedents).

Instead of LSA, the strongest performers here were the BERT models which produced the best micro- and macro-recall on the 10\% and 50\% subsets and $glove_{cnn}$ for the 100\% training subset.

\section{Discussion}

We initially expected pre-trained language models, being the state-of-the-art on many non-legal NLP tasks, to perform best here as well. That an LSA-based linSVM would outperform both word-embedding and language models by many measures surprised us. How LSA achieved this is explored in Appendix \ref{app:topic_top_words} which presents a sample of the (quite informative) topics extracted. %Considering also that a keyword-rules-based model produced strong recall, there is an argument for applying Occam's razor here.

One caveat to interpreting our results: we focused on comparing the models' \textit{out-of-box} performance, rather than comparing the models at their best (i.e.\ after extensive cross-validation and tuning). Specifically, the BERT models' inability to be fine-tuned on longer input texts meant that they competed at a disadvantage, having been shown only selected judgment portions. Despite this, BERT models proved competitive on smaller training subsets. Likewise, while $ulmfit$ performed well on the 10\% subset (suggesting that it benefited from encoder pre-training), the model struggled to leverage additional training data and recorded only modest improvements on the larger training subsets.

Thus, our answer to the research question stated in Section \ref{sec:intro} is a nuanced one: while state-of-the-art models do not clearly outperform traditional statistical models when applied \textit{out-of-box} to legal corpora, they show promise for dealing with data constraints particularly if further adapted and fine-tuned to accommodate longer texts. This should inform future research.

\section{Conclusion and Future Work}

This paper comparatively benchmarked traditional topic models against more recent, sophisticated, and computationally intensive techniques on the legal area classification task. We found that while data scarcity affects all ML classifiers, certain classifiers, especially pre-trained language models, could perform well with as few as 588 labelled judgments.

Our results also suggest that more work can be done to adapt state-of-the-art NLP models for the legal domain. Two areas seem promising: (1) creating \textit{law-specific} datasets and baselines for training and benchmarking legal text classifiers, and (2) exploring representation learning techniques that leverage transfer learning methods but scale well on long texts. For the latter, possible directions here include exploring different CNN architectures and their hyperparameters, using contextualized word embeddings, and using feature extraction methods on pre-trained language models like BERT (as opposed to fine-tuning them) so that they can be used on longer text inputs. As Lord Denning said in \textit{Packer v Packer} [1953] EWCA Civ J0511-3:

\begin{quote}
``If we never do anything which has not been done before, we shall never get anywhere. The law will stand whilst the rest of the world goes on; and that will be bad for both.''
\end{quote}

\section*{Acknowledgments}

We thank the anonymous reviewers for their helpful comments and the Singapore Academy of Law for permitting us to scrape and use this corpus.

\bibliography{naaclhlt2019}
\bibliographystyle{acl_natbib}

\appendix

% \section{Appendices}
% \label{sec:appendix}
% Appendices are material that can be read, and include lemmas, formulas, proofs, and tables that are not critical to the reading and understanding of the paper. 
% Appendices should be {\bf uploaded as supplementary material} when submitting the paper for review. Upon acceptance, the appendices come after the references, as shown here. Use
% \verb|\appendix| before any appendix section to switch the section
% numbering over to letters.

\section{Appendices}
\label{sec:supplemental}

\subsection{Data Parsing}

The original scraped dataset had 6,839 judgments in PDF format. The PDFs were parsed with a custom Python script using the \texttt{pdfplumber}\footnote{\url{https://github.com/jsvine/pdfplumber}} library. For the purposes of our experiments, we excluded the case information section found at the beginning of each PDF as we did not consider it to be part of the judgment (this section contains information such as Case Number, Decision Date, Coram, Counsel Names etc). The labels were extracted based on their location in the first page of the PDF, i.e. immediately after the case information section and before the author line. After this process, 611 judgments that were originally unlabelled and one incorrectly parsed judgment were dropped, leaving the final dataset of 6,227 judgments.

\subsection{Label Mappings}
\label{app:label_map}

Labels are a double-dash-delimited series of increasingly specific legal $N$-grams (e.g.\ ``tort--negligence--duty of care--whether occupier owes lawful entrants a duty of care''), which denote increasingly specific and narrow areas of law. Multiple labels are expressed in multiple lines (one label per line). We checked the topic labels for consistency and typographical errors by inspecting a list of unique labels across the dataset. Erroneous labels and labels that were conceptual subsets of others were manually mapped to primary labels via the mapping presented in Table \ref{label_map}. Some subjectivity admittedly exists in the choice of mappings. However we were not aware of any standard ontology for used for legal area classification, particularly for Singapore law. To mitigate this, we based the primary label set on the Singapore Academy of Law Subject Tree which, for copyright reasons, we were unable to reproduce here.

\begin{table*}[p!]
\resizebox{\textwidth}{!}{%
\begin{tabular}{lp{12cm}}
\toprule
                                        Primary Label &                                                                                                                                       Alternative Labels \\
\midrule
              administrative\_and\_constitutional\_law &                                              administrative\_law, adminstrative\_law, constitutional\_interpretation, constitutional\_law, elections \\
                admiralty\_shipping\_and\_aviation\_law &                                                                             admiralty, admiralty\_and\_shipping, carriage\_of\_goods\_by\_air\_and\_land \\
                                         agency\_law &                                                                                                                                           agency \\
                                        arbitration &                                                                                                                                                  \\
                                        banking\_law &                                                                                                                                          banking \\
                            biomedic\_law\_and\_ethics &                                                                                                                                                  \\
                      building\_and\_construction\_law &                                                                                                              building\_and\_construction\_contracts \\
                                    civil\_procedure &                                       civil\_procedue, application\_for\_summary\_judgment, limitation\_of\_actions, procedure, discovery\_of\_documents \\
                                        company\_law &                                                                                           companies, companies-\_meetings, companies\_-\_winding\_up \\
                                    competition\_law &                                                                                                                                                  \\
                                   conflict\_of\_laws &                                                                                                              conflicts\_of\_laws, conflicts\_of\_law \\
                                       contract\_law &                                                            commercial\_transactions, contract, contract\_-\_interpretation, contracts, transactions \\
                                       criminal\_law &                                                                                                                contempt\_of\_court, offences, rape \\
                  criminal\_procedure\_and\_sentencing &                                                                                        criminal\_procedure, criminal\_sentencing, sentencing, bail \\
                                credit\_and\_security &                                                                                                         credit\_and\_securities, credit\_\&\_security \\
                                            damages &                                                                                            damage, damages\_-\_assessment, injunction, injunctions \\
                                           evidence &                                                                                                                                     evidence\_law \\
                                     employment\_law &                                                                                                                     work\_injury\_compensation\_act \\
                                  equity\_and\_trusts &                                                                                                                equity, estoppel, trusts, tracing \\
                                         family\_law &                                                                       succession\_and\_wills, probate\_\&\_administration, probate\_and\_administration \\
                                     insolvency\_law &                                                                                                                                       insolvency \\
                                      insurance\_law &                                                                                                                                        insurance \\
                          intellectual\_property\_law &  intellectual\_property, copyright, copyright\_infringement, designs, trade\_marks\_and\_trade\_names, trade\_marks, trademarks, patents\_and\_inventions \\
                                  international\_law &                                                                                                                                                  \\
                              non\_land\_property\_law &                                                                                                personal\_property, property\_law, choses\_in\_action \\
                                           land\_law &                                                                                                          landlord\_and\_tenant, land, planning\_law \\
                                   legal\_profession &                                                                                                                               legal\_professional \\
                                         muslim\_law &                                                                                                                                                  \\
                                    partnership\_law &                                                                                                                        partnership, partnerships \\
                                        restitution &                                                                                                                                                  \\
                                revenue\_and\_tax\_law &                                                                                                                        tax, revenue\_law, tax\_law \\
                                           tort\_law &                                                                                                                           tort, abuse\_of\_process \\
                                  words\_and\_phrases &                                                                                                                         statutory\_interpretation \\
                                       res\_judicata &                                                                                                                                                  \\
                                        immigration &                                                                                                                                                  \\
                            courts\_and\_jurisdiction &                                                                                                                                                  \\
                                       road\_traffic &                                                                                                                                                  \\
                                  debt\_and\_recovery &                                                                                                                                                  \\
                                           bailment &                                                                                                                                                  \\
                                          charities &                                                                                                                                                  \\
       unincorporated\_associations\_and\_trade\_unions &                                                                                                                      unincorporated\_associations \\
                                        professions &                                                                                                                                                  \\
 bills\_of\_exchange\_and\_other\_negotiable\_instruments &                                                                                                                                                  \\
                                              gifts &                                                                                                                                                  \\
                     mental\_disorders\_and\_treatment &                                                                                                                                                  \\
                        deeds\_and\_other\_instruments &                                                                                                                                                  \\
                   financial\_and\_securities\_markets &                                                                                                                                                  \\
                              sheriffs\_and\_bailiffs &                                                                                                                                                  \\
                                            betting &                                                                                                      \_gaming\_and\_lotteries, gaming\_and\_lotteries \\
                                      sale\_of\_goods &                                                                                                                                                  \\
                                               time &                                                                                                                                                  \\
\bottomrule
\end{tabular}%
}
\caption{Primary-Alternative mappings for raw dataset labels}
\label{label_map} 
\end{table*}

\begin{figure*}[p!]
    \includegraphics[width=13.5cm]{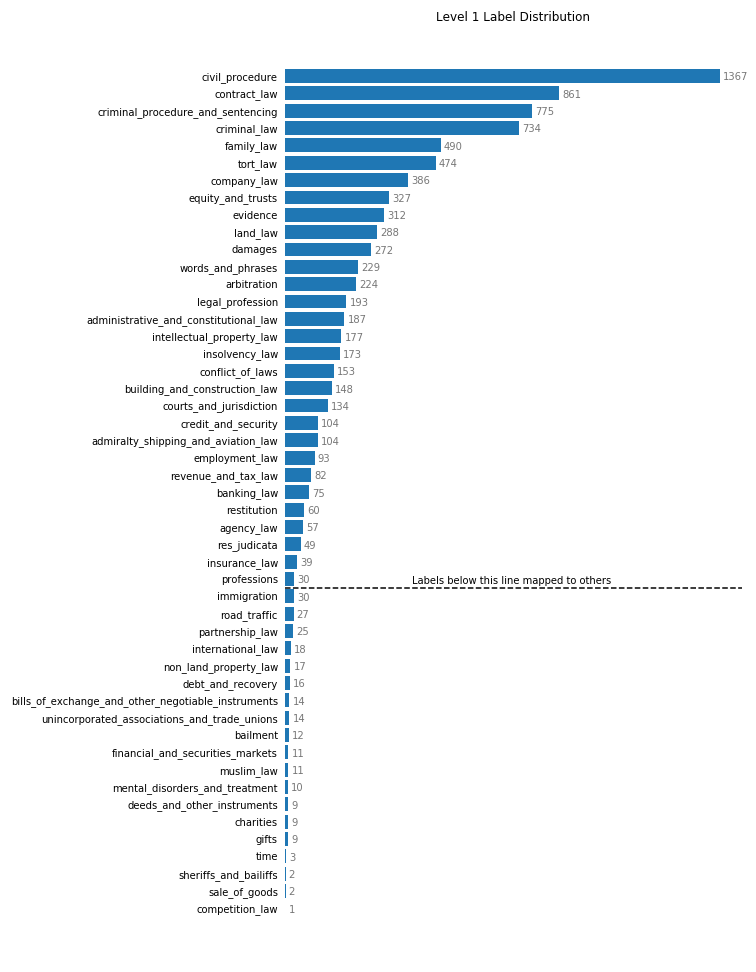}
    \caption{Distribution of Cleaned Labels and the Final 30 Labels Included}
    \label{fig:label_dist}
\end{figure*}

It was only \textit{after} this step that the top 30 labels were kept and the remaining mapped to ``others''. Figure \ref{fig:label_dist} presents all 51 original labels and their frequencies.

\subsection{Implementation Details on Models Used}
\label{app:implementation}

All text preprocessing (tokenization, stopping, and lemmatization) was done using \texttt{spaCy} defaults \cite{spacy2}.

\subsubsection{Baseline Models}

$count_m$ uses the FlashText algorithm for efficient exact phrase matching within long judgment texts \cite{flashtext}. To populate the set of associated terms for each label, all sub-labels attributable to the label within the given training subset were first added as exact phrases. Next, the label itself was tokenized into unigrams. Each unigram was added individually to the set of associated terms unless it fell within a set of customized stopwords we created after inspecting all labels. The set is $\{and, law, of, non, others\}$.

Beyond $count_{25}$, we experimented with thresholds of 1, 5, 10, and 35 occurrences. F1 scores increased linearly as thresholds increased from 1 to 25, but only increased marginally from 25 to 35.

\subsubsection{Topic Models}

LSA was achieved using scikit-learn's TFIDFVectorizer and TruncatedSVD classes. Document-topic weights were then normalized with scikit-learn's Normalizer class before being fed to the classifier. Where relevant, the random state was set at 36. Note that judgments were preprocessed with spaCy as above before being fed into the LSA pipeline. Beyond 100 and 250 topics, an experiment using 50 topics only performed consistently worse.

The classifier used scikit-learn's OneVsRest and LinearSVC classes with all default settings. An alternative linSVM with balanced class-weights was tested but performed consistently worse by both macro and micro-f1 scores and was thus omitted for brevity.

\subsubsection{Word Embedding Feature Models}

For all the word embedding feature models, we used spaCy's tokenizer to obtain the word tokens. We fixed the maximum sequence length per judgment at 10K tokens and used a vocabulary of the top 60K most common words in the training corpus. Words that did not have a corresponding GloVe vector were initialized from a uniform distribution with range $[-0.5, 0.5]$. The models were implemented in TensorFlow\footnote{\url{https://github.com/tensorflow/tensorflow}} with the Keras API. To deal with class imbalance, we weighted the losses by passing class weights to the \texttt{class\_weight} argument of \texttt{model.fit}.

For the CNN models, we based our implementation off the non-static version in \citet{kim2014emnlp} but used [3, 3, 3] x 600 filters, as we found that increasing the number of filters improved results.

\subsubsection{BERT}

To fine-tune BERT to our multi-label classification task, we used the PyTorch implementation of BERT by \texttt{HuggingFace}\footnote{\url{https://github.com/huggingface/pytorch-pretrained-BERT}} and added a linear classification layer $W \in \mathbf{R}^{K \times H}$, where $K$ is the number of classifier labels and $H$ is the dimension of the pooled representation of the input sequence, followed by a sigmoid function. We fine-tuned all the BERT models using mixed-precision training and gradient accumulation (8 steps). To address data imbalance, we weighted the losses by passing positive weights for each class (capped at 30) to the \texttt{pos\_weight} argument of \texttt{torch.nn.BCEWithLogitsLoss}. 

\subsubsection{ULMFiT}

We first fine-tuned the pre-trained ULMFiT language model (WikiText-103) on our entire corpus of 6,227 judgments using a language model objective for 10 epochs before replacing the output layer with a classifier output layer and then further fine-tuned the model on labelled data with the classification objective using \texttt{fastai}'s recommended recipe\footnote{\url{https://docs.fast.ai/text.html}} for text classification (we used gradual unfreezing and the one-cycle learning rate schedule to fine-tune the classifier until there was no more improvement on the validation score). We used mixed precision training and fixed the maximum sequence length at 5K tokens to allow the training data to fit in memory.

\subsection{Topics Extracted by Topic Mining}
\label{app:topic_top_words}

Table \ref{topic_words} presents the top 10 tokens associated with the top 25 topics extracted by LSA on the 100\% data subset. Notice that these topics are common to both $lsa_{100}$ and $lsa_{250}$ since the output of TFIDF and SVD do not vary with $k$. The only difference is that $lsa_{100}$ uses only the first 100 topic vectors (i.e.\ the topic vectors corresponding to the 100 largest singular values computed by the decomposition) created by LSA whereas $lsa_{250}$ uses the first 250. However, topics extracted from different data subsets would differ.

A quick perusal of the extract topics suggests many have would be highly informative of a case's legal area. Topics 2, 7, 21, 24, and 25 map nicely to criminal law, topics 3 and 5 to family law, and topics 18, and 20 to arbitration. Other individually-informative topics include topics 6 (road traffic), 8 (building and construction law), 9 (land law), 11 (legal profession), 16 (company law), and 22 (conflict of laws).

\begin{table*}[p!]
\begin{center}
\resizebox{\textwidth}{!}{
\begin{tabular}{ll}
\toprule
Topic No.\ & Top 10 Tokens \\
\midrule
1  &  plaintiff, court, defendant, case, party, claim, order, appeal, fact, time \\
2  &  offence, accuse, sentence, imprisonment, prosecution, offender, charge, drug, convict, conviction \\
3  &  matrimonial, husband, wife, marriage, child, maintenance, contribution, asset, cpf, divorce \\
4  &  application, court, appeal, order, district, matrimonial, respondent, proceeding, judge, file \\
5  &  matrimonial, marriage, child, maintenance, husband, divorce, parliament, division, context, broad \\
6  &  injury, accident, plaintiff, damage, defendant, dr, award, medical, work, pain \\
7  &  drug, cnb, diamorphine, packet, mda, bag, heroin, traffic, arbitration, plastic \\
8  &  contractor, contract, sentence, imprisonment, construction, clause, project, offender, cl, payment \\
9  &  property, land, purchaser, tenant, title, estate, decease, owner, road, lease \\
10 &  arbitration, victim, rape, sexual, arbitrator, arbitral, cane, clause, accuse, cl \\
11 &  disciplinary, profession, solicitor, committee, advocate, society, client, misconduct, lpa, professional \\
12 &  creditor, debt, bankruptcy, accident, debtor, wind, liquidator, injury, death, decease \\
13 &  plaintiff, defendant, proprietor, infringement, plaintiffs, defendants, cane, 2014, land, 2012 \\
14 &  appellant, 2014, road, district, 2016, trial, defendant, property, judge, pp \\
15 &  drug, arbitration, profession, disciplinary, society, clause, vessel, death, advocate, diamorphine \\
16 &  shareholder, director, company, vehicle, share, resolution, traffic, management, vote, minority \\
17 &  creditor, solicitor, road, vehicle, profession, bankruptcy, disciplinary, drive, lane, driver \\
18 &  arbitration, adjudicator, decease, tribunal, adjudication, arbitral, vehicle, arbitrator, drive, mark \\
19 &  contractor, adjudicator, adjudication, decease, beneficiary, estate, employer, death, child, executor \\
20 &  arbitration, arbitrator, tribunal, award, arbitral, profession, contractor, disciplinary, architect, lpa \\
21 &  drug, respondent, appellant, diamorphine, gd, factor, cl, adjudicator, judge, creditor \\
22 &  2015, forum, 2014, 2016, 2013, foreign, 2012, appellant, conveniens, spiliada \\
23 &  stay, appellant, arbitration, estate, register, forum, district, beneficiary, owner, applicant \\
24 &  vessel, cargo, decease, murder, sale, ship, death, dr, kill, knife \\
25 &  sexual, rape, penis, vagina, complainant, stroke, intercourse, penetration, vessel, sex \\
\bottomrule
\end{tabular}}
\end{center}
\caption{\label{topic_words} Top Tokens For Top 25 Topics Extracted by $lsa_{250}$ on the 100\% subset.}
\end{table*}

% \subsection{Template instructions -- REMOVE BEFORE SUBMISSION}

% Submissions may include non-readable supplementary material used in the work and described in the paper. Any accompanying software and/or data should include licenses and documentation of research review as appropriate. Supplementary material may report preprocessing decisions, model parameters, and other details necessary for the replication of the experiments reported in the paper. Seemingly small preprocessing decisions can sometimes make a large difference in performance, so it is crucial to record such decisions to precisely characterize state-of-the-art methods. 

% Nonetheless, supplementary material should be supplementary (rather
% than central) to the paper. {\bf Submissions that misuse the supplementary 
% material may be rejected without review.}
% Supplementary material may include explanations or details
% of proofs or derivations that do not fit into the paper, lists of
% features or feature templates, sample inputs and outputs for a system,
% pseudo-code or source code, and data. (Source code and data should
% be separate uploads, rather than part of the paper).

% The paper should not rely on the supplementary material: while the paper
% may refer to and cite the supplementary material and the supplementary material will be available to the reviewers, they will not be asked to review the
% supplementary material.

\end{document}